\documentclass[11pt,a4paper]{article}
\usepackage[hyperref]{emnlp2020}
\usepackage{times}
\usepackage{latexsym}

\usepackage{microtype}
\usepackage{graphicx}
\usepackage{subfigure}
\usepackage{booktabs} 
\usepackage{hyperref}
\usepackage{url}
\usepackage{multirow}
\usepackage{wrapfig}
\usepackage{enumitem}
\usepackage{times}
\usepackage{latexsym}
\usepackage{amsmath}
\usepackage{amsthm}
\usepackage{amssymb}
\usepackage{xspace}
\usepackage{amsfonts,eucal,amsbsy}
\usepackage[normalem]{ulem}
\usepackage{sidecap}
\sidecaptionvpos{figure}{t}

\usepackage{amsmath,amsfonts,bm}

\def\eqref#1{equation~\ref{#1}}

\def\1{\bm{1}}

\DeclareMathAlphabet{\mathsfit}{\encodingdefault}{\sfdefault}{m}{sl}
\SetMathAlphabet{\mathsfit}{bold}{\encodingdefault}{\sfdefault}{bx}{n}

\newcommand{\E}{\mathbb{E}}

\newcommand{\spavg}{\textsc{SP}\xspace}
\newcommand{\lstmavg}{\textsc{BiLSTM}\xspace}
\newcommand{\monoae}{\textsc{EnglishAE}\xspace}
\newcommand{\monovae}{\textsc{EnglishVAE}\xspace}
\newcommand{\engtrans}{\textsc{EnglishTrans}\xspace}
\newcommand{\multitrans}{\textsc{BilingualTrans}\xspace}
\newcommand{\multitransprior}{\textsc{BGT w/o LangVars}\xspace}
\newcommand{\trans}{\textsc{BGT w/o Prior}\xspace}
\newcommand{\vae}{\textsc{BGT}\xspace}

\usepackage{microtype}

\aclfinalcopy 

\title{A Bilingual Generative Transformer for Semantic Sentence Embedding}

\author{John Wieting$^1$, Graham Neubig$^1$, and Taylor Berg-Kirkpatrick$^2$ \\
  $^1$Carnegie Mellon University,
  Pittsburgh, PA, 15213, USA \\
  $^2$University of California San Diego,
  San Diego, CA, 92093, USA\\
  {\small \texttt{\{jwieting,gneubig\}@cs.cmu.edu}, \texttt{tberg@eng.ucsd.edu}}}

\date{}

\begin{document}
\maketitle
\begin{abstract}
Semantic sentence embedding models encode natural language sentences into vectors, such that closeness in embedding space indicates closeness in the semantics between the sentences. 
Bilingual data offers a useful signal for learning such embeddings: properties shared by both sentences in a translation pair are likely semantic, while divergent properties are likely stylistic or language-specific. We propose a deep latent variable model that attempts to perform source separation on parallel sentences, isolating what they have in common in a latent semantic vector, and explaining what is left over with language-specific latent vectors. Our proposed approach differs from past work on semantic sentence encoding in two ways. First, by using a variational probabilistic framework, we introduce priors that encourage source separation, and can use our model's posterior to predict sentence embeddings for monolingual data at test time. Second, we use high-capacity transformers as both data generating distributions and inference networks -- contrasting with most past work on sentence embeddings.
In experiments, our approach substantially outperforms the state-of-the-art on a standard suite of unsupervised
semantic similarity evaluations.
Further, we demonstrate that our approach yields the largest gains on more difficult subsets of these evaluations where simple word overlap is not a good indicator of similarity.
\footnote{Code and data to replicate results available at \url{https://www.cs.cmu.edu/~jwieting}.}
\end{abstract}

\section{Introduction}
Learning useful representations of language has been a source of recent success in natural language processing (NLP).
Much work has been done on learning representations for  words~\citep{mikolov2013distributed, pennington2014glove} and sentences~\citep{kiros2015skip, conneau2017supervised}. More recently, deep neural architectures have been used to learn contextualized word embeddings~\citep{peters2018deep, devlin2018bert} enabling state-of-the-art results on many tasks. We focus on learning semantic \emph{sentence} embeddings in this paper, which play an important role in many downstream applications. Since they do not require any labelled data for fine-tuning, sentence embeddings are useful out-of-the-box for problems such as measurement of Semantic Textual Similarity (STS; \citet{agirre2012semeval}), mining bitext \citep{zweigenbaum2018overview}, and paraphrase identification \citep{dolan-04}. Semantic similarity measures also have downstream uses such as fine-tuning machine translation systems~\citep{wieting2019beyond}.

There are three main ingredients when designing a sentence embedding model: the architecture, the training data, and the  objective function.
Many architectures including 
LSTMs~\citep{hill2016learning,conneau2017supervised,schwenk2017learning,subramanian2018learning}, Transformers~\citep{cer2018universal,reimers2019sentence}, and averaging models~\citep{wieting-16-full,arora2017simple} are capable of learning sentence embeddings. The choice of training data and objective are intimately intertwined, and there are a wide variety of options including next-sentence prediction~\citep{kiros2015skip}, machine translation~\citep{espana2017empirical,schwenk2017learning,schwenk2018filtering,artetxe2018massively}, natural language inference (NLI) ~\citep{conneau2017supervised}, and multi-task objectives which include some of the previously mentioned objectives~\citep{cer2018universal} potentially combined with additional tasks like constituency parsing~\citep{subramanian2018learning}.

Surprisingly, despite ample testing of more powerful architectures, the best performing models for many sentence embedding tasks related to semantic similarity often use simple architectures that are mostly agnostic to the interactions between words.
For instance, some of the top performing techniques use word embedding averaging~\citep{wieting-16-full}, character n-grams~\citep{wieting2016charagram}, and subword embedding averaging~\citep{wieting2019para} to create representations. These simple approaches are competitive with much more complicated architectures on in-domain data and generalize well to unseen domains, but are fundamentally limited by their inability to capture word order. 
Training these approaches generally relies on discriminative objectives defined on paraphrase data~\citep{GanitkevitchDC13-short,wieting2017pushing}
or bilingual data~\citep{wieting2019para,chidambaram-etal-2019-learning,yang-etal-2020-multilingual}. The inclusion of latent variables in these models has also been explored~\citep{chen2019multi}.

Intuitively, bilingual data in particular is promising because it potentially offers a useful signal for learning the underlying semantics of sentences.
Within a translation pair, properties shared by both sentences are more likely semantic, while those that are divergent are more likely stylistic or language-specific.
While previous work learning from bilingual data perhaps takes advantage of this fact \emph{implicitly}, the focus of this paper is modelling this intuition \emph{explicitly}, and to the best of our knowledge, this has not been explored in prior work.
Specifically, we propose a deep generative model that is encouraged to perform {\it source separation} on parallel sentences, isolating what they have in common in a latent \emph{semantic embedding} and explaining what is left over with \emph{language-specific latent vectors}.
At test time, we use inference networks~\citep{kingma2013auto} for approximating the model's posterior 
on the semantic and source-separated latent variables
to encode monolingual sentences.
Finally, since our model and training objective are generative, 
our approach does not require knowledge of the distance metrics to be used during evaluation,
and it has the additional property of being able to generate text.

In experiments, we evaluate our probabilistic source-separation approach 
on a standard suite of
STS evaluations.
We demonstrate that the proposed approach is effective, most notably allowing the learning of high-capacity deep Transformer architectures~\citep{vaswani2017attention} while still generalizing to new domains, significantly outperforming a variety of state-of-the-art baselines.
Further, we conduct a thorough analysis by identifying subsets of the STS evaluation where simple word overlap is not able to accurately assess semantic similarity.
On these most difficult instances, we find that our approach yields the largest gains, indicating that our system is modeling interactions between words 
to good effect. We also find that our model better handles cross-lingual semantic similarity than multilingual translation baseline approaches, indicating that stripping away language-specific information allows for better comparisons between sentences from different languages.

Finally, we analyze our model to uncover what information was captured by the source separation into the semantic and language-specific variables and the relationship between this encoded information and language distance to English. We find that the language-specific variables tend to explain more superficial or language-specific properties such as overall sentence length, amount and location of punctuation, and the gender of articles (if gender is present in the language), but semantic and syntactic information is more concentrated in the shared semantic variables, matching our intuition. Language distance has an effect as well, where languages that share common structures with English put more information into the semantic variables, while more distant languages put more information into the language-specific variables.
Lastly, we show outputs generated from our model that exhibit its ability to do a type of style transfer.

\section{Model} \label{sec:models}
\begin{figure}[t]
    \centering
    \small
    \includegraphics[scale=0.24, trim = {0cm 0cm 0cm 0cm}]{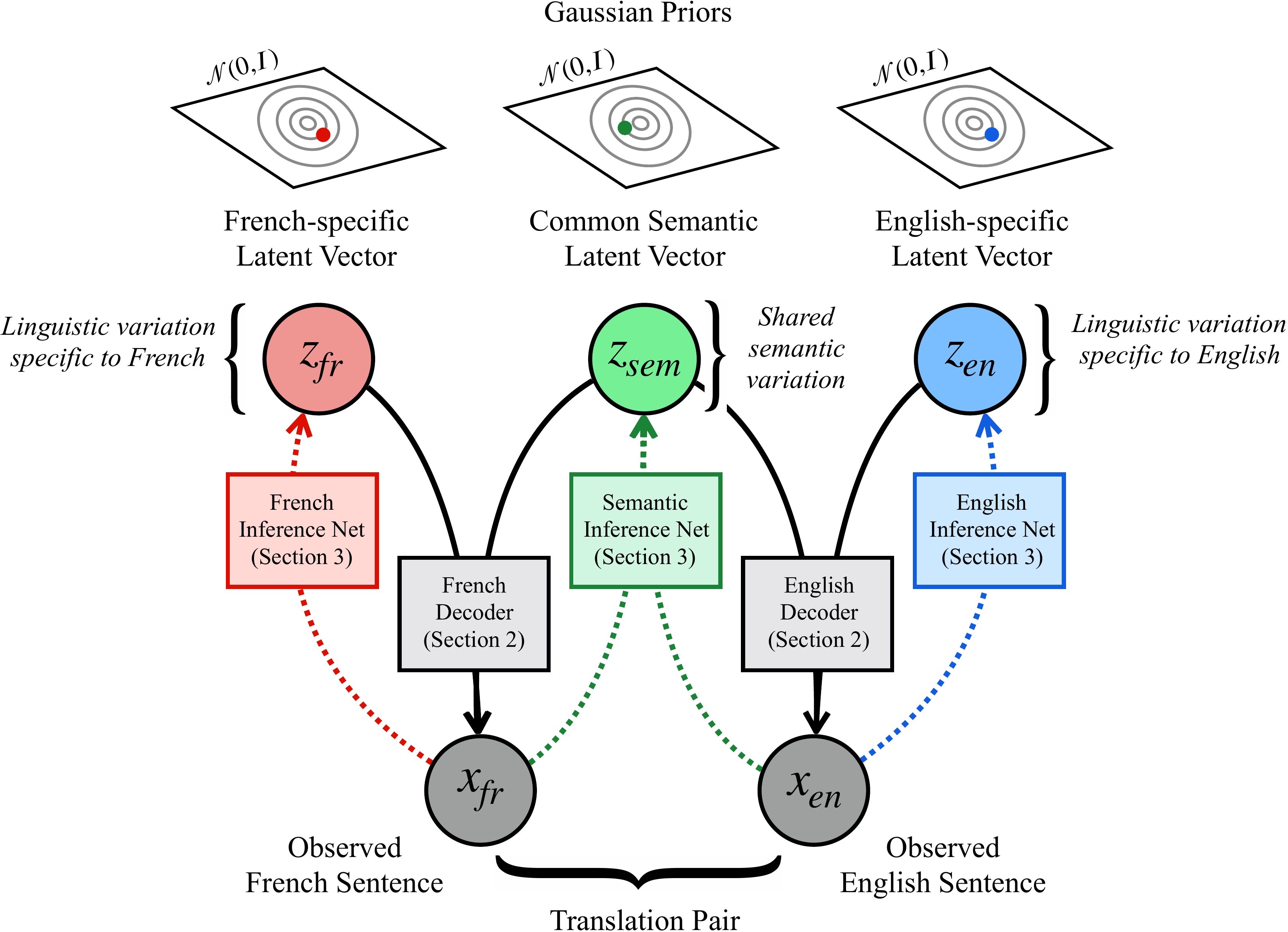}
    \caption{The generative process of our model. Latent variables modeling the linguistic variation in French and English, $z_{fr}$ and $z_{en}$, as well as a latent variable modeling the common semantics, $z_{sem}$, are drawn from a multivariate Gaussian prior. The observed text in each language is then conditioned on its language-specific variable and $z_{sem}$. 
    }
    \label{fig:generative}
\end{figure}

\begin{figure*}
    \centering
    \small
    \includegraphics[scale=0.32, trim = {0cm 0cm 0cm 0cm}]{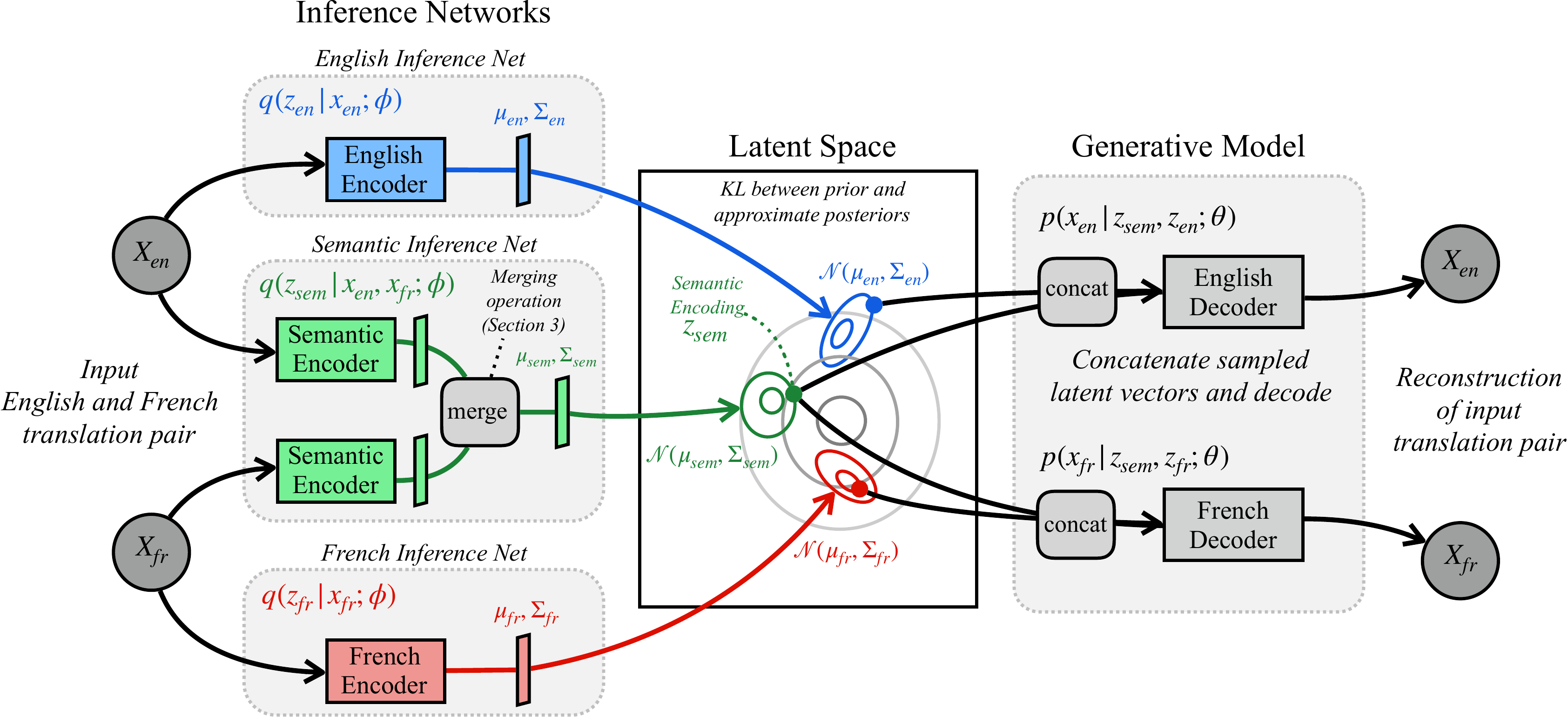}
    \caption{The computation graph for the variational lower bound used during training. The English and French text are fed into their respective inference networks and the semantic inference network to ultimately produce the language variables $z_{fr}$ and $z_{en}$ and semantic variable $z_{sem}$. Each language-specific variable is then concatenated to $z_{sem}$ and used by the decoder to reconstruct the input sentence pair.
    }
    \label{fig:computationgraph}
\end{figure*}

Our proposed training objective leverages a generative model of parallel text in two languages. Our model is language agnostic, and applies to a wide variety of languages (see Section~\ref{sec:analysis}) but we will use the running example of English (\texttt{en}) and French (\texttt{fr}) pairs consisting of an English sentence $x_{en}$ and a French sentence $x_{fr}$. Importantly, the generative process utilizes three underlying latent vectors: language-specific variation variables (language variables) $z_{fr}$ and $z_{en}$ for each side of the translation, as well as a shared semantic variation variable (semantic variable) $z_{sem}$.
In this section we will first describe the generative model for the text and latent variables.
In the following section we will describe the inference procedure of $z_{sem}$ given an input sentence, which corresponds to our core task of obtaining sentence embeddings useful for downstream tasks such as semantic similarity.

The generative process of our model, the Bilingual Generative Transformer (\vae), is depicted in Figure~\ref{fig:generative} and the training computation graph is shown in Figure~\ref{fig:computationgraph}.
First, we sample latent variables $\langle z_{fr}, z_{en}, z_{sem} \rangle$, where $z_i \in \mathbb{R}^k$, from a multivariate Gaussian prior $\mathcal{N}(0,I_k)$.
These variables are then fed into a decoder that samples sentences; $x_{en}$ is sampled conditioned on $z_{sem}$ and $z_{en}$, while $x_{fr}$ is sampled conditioned on $z_{sem}$ and $z_{fr}$.
Because sentences in both languages will use $z_{sem}$ in generation, we expect that in a well-trained model this variable will encode semantic, syntactic, or stylistic information shared across both sentences, while $z_{fr}$ and $z_{en}$ will handle any language-specific peculiarities or specific stylistic decisions that are less central to the sentence meaning and thus do not translate across sentences.
In the following section, we further discuss how this is explicitly encouraged by the learning process.

\noindent
\textbf{Decoder Architecture.}
Many latent variable models for text use LSTMs~\citep{hochreiter1997long} as their decoders~\citep{yang2017improved,ziegler2019latent,ma2019flowseq}.
However, state-of-the-art models in neural machine translation have seen increased performance and speed using deep Transformer architectures. We also found in our experiments (see Appendix~\ref{appendix:ablations} for details) that Transformers led to increased performance in our setting, so they are used in our main model. 

We use two decoders in our model, one for modelling $p(x_{fr}|z_{sem}, z_{fr};\theta)$ and one for modeling $p(x_{en}|z_{sem}, z_{en};\theta)$ (see right side of Figure~\ref{fig:computationgraph}). Each decoder takes in a language variable and a semantic variable, which are concatenated and used by the decoder for reconstruction. We explore four ways of using this latent vector:
(1) Concatenate it to the word embeddings (Word) (2) Use it as the initial hidden state (Hidden, LSTM only) (3) Use it as you would the  attention context vector in the traditional sequence-to-sequence framework (Attention) and (4) Concatenate it to the hidden state immediately prior to computing the logits (Logit). Unlike Attention, there is no additional feedforward layer in this setting.
We experimented with these four approaches, as well as combinations thereof, and report this analysis in Appendix~\ref{appendix:embedding}. From these experiments, we see that the closer the sentence embedding is to the final word predictor, the better the performance on downstream tasks evaluating its semantic content. We hypothesise that this is due to better gradient propagation because the sentence embedding is now closer to the error signal. 
Since Attention and Logit performed best, we use these in our main experiments.

\section{Learning and Inference} \label{sec:learning}

Our model is trained on a set of parallel sentences $X$ consisting of $N$ examples, $X = \{\langle x^1_{en}, x^1_{fr} \rangle, \ldots, \langle x^N_{en}, x^N_{fr} \rangle\}$, and $Z$ is our collection of latent variables $Z=(\langle z^1_{en}, z^1_{fr}, z^1_{sem} \rangle, \ldots, \langle z^N_{en}, z^N_{fr}, z^N_{sem} \rangle)$.
We wish to maximize the likelihood of the parameters of the two decoders $\theta$ with respect to the observed $X$, marginalizing over the latent variables $Z$.

$$
p(X;\theta) = \int_Z p(X,Z; \theta)dZ
$$

Unfortunately, this integral is intractable due to the complex relationship between $X$ and $Z$.
However, related latent variable models like variational autoencoders (VAEs; \citet{kingma2013auto}) learn by optimizing a variational lower bound on the log
marginal likelihood. This surrogate objective is called the evidence lower bound (ELBO) and introduces a variational approximation, $q$ to the true posterior of the model $p$. The $q$ distribution is parameterized by a neural network with parameters $\phi$. ELBO for our model is written as:

\begin{align*}
\text{ELBO} = &\E_{q(Z|X;\phi)} [\log p(X|Z;\theta)] -\\ 
       &\text{KL}(q(Z|X;\phi) || p(Z;\theta))
\end{align*}

This lower bound on the marginal can be optimized by gradient ascent by using the reparameterization trick~\citep{kingma2013auto}. This trick allows for the expectation under $q$ to be approximated through sampling in a way that preserves backpropagation.
We make several independence assumptions for $q(z_{sem},z_{en},z_{fr}|x_{en}, x_{fr};\phi)$ to match our goal of source separation: we factor $q$ as $q(z_{sem} | x_{en}, x_{fr};\phi) q(z_{en} | x_{en};\phi) q(z_{fr} | x_{fr};\phi)$.
The parameters of the encoders that make up the inference networks, defined in the next paragraph, are denoted as $\phi$.

Lastly, we note that the KL term in our ELBO equation \emph{explicitly encourages} explaining variation that is shared by translations with the shared semantic variable, and explaining language-specific variation with the corresponding language-specific variables.
Encoding information shared by the two sentences in the shared variable results in only a single penalty from the KL loss, while encoding the information separately in both language specific variables will cause unnecessary replication, doubling the overall cost incurred by the KL term.

\paragraph{Encoder Architecture.}

We use three inference networks as shown on the left side of  Figure~\ref{fig:computationgraph}: an English inference network to produce the English language variable, a French inference network to produce the French language variable, and a semantic inference network to produce the semantic variable. Just as in the decoder architecture, we use a Transformer for the encoders. 

The semantic inference network is a bilingual encoder that encodes each language. For each translation pair, we alternate which of the two parallel sentences is fed into the semantic encoder within a batch. Since the semantic encoder is meant to capture language agnostic semantic information, its outputs for a translation pair should be similar regardless of the language of the input sentence. We note that other operations are possible for combining the views each parallel sentence offers. For instance, we could feed both sentences into the semantic encoder and pool their representations. However, in practice we find that alternating works well and also can be used to obtain sentence embeddings for text that is not part of a translation pair. We leave further study of combining views to future work.

\section{Experiments}
\subsection{Baseline Models} \label{sec:baseline}
We experiment with fourteen baseline models, covering both the most effective approaches for learning sentence embeddings from the literature and ablations of our own \vae model. These baselines can be split into three groups as detailed below.

\paragraph{Models from the Literature (Trained on Different Data)} We compare to well known sentence embedding models Infersent~\citep{conneau2017supervised}, GenSen~\citep{subramanian2018learning}, the Universal Sentence Encoder (USE)~\citep{cer2018universal}, LASER~\cite{artetxe2018massively}, as well as BERT~\citep{devlin2018bert}.\footnote{
As shown in Table~\ref{table:results}, BERT doesn't perform exceptionally well on STS tasks. We found the same to be true for the masked language models XLM~\cite{conneau2019cross} and XLM-R~\cite{conneau2019unsupervised} for both monolingual and cross-lingual STS tasks.
}\footnote{Note that in all experiments using BERT, including Sentence-BERT, the large, uncased version is used.} We used the pretrained BERT model in two ways to create a sentence embedding. The first way is to concatenate the hidden states for the CLS token in the last four layers. The second way is to concatenate the hidden states of all word tokens in the last four layers and mean pool these representations. Both methods result in a 4096 dimension embedding. We also compare to the newly released model, Sentence-Bert~\citep{reimers2019sentence}. This model is similar to Infersent~\citep{conneau2017supervised} in that it is trained on natural language inference data, SNLI~\citep{bowman2015large}. However, instead of using pretrained word embeddings, they fine-tune BERT in a way to induce sentence embeddings.\footnote{Most work evaluating accuracy on STS tasks has averaged the Pearson's $r$ over each individual dataset for each year of the STS competition. However, \citet{reimers2019sentence} computed Spearman's $\rho$ over concatenated datasets for each year of the STS competition. To be consistent with previous work, we re-ran their model and calculated results using the standard method, and thus our results are not the same as those reported \citet{reimers2019sentence}.}

\paragraph{Models from the Literature (Trained on Our Data)} These models are amenable to being trained in the exact same setting as our own models as they only require parallel text. These include the sentence piece averaging model, \spavg, from~\citet{wieting2019para}, which is among the best of the averaging models (i.e. compared to averaging only words or character $n$-grams) as well the LSTM model, \lstmavg, from~\citet{wieting-17-full}. These models use a contrastive loss with a margin. Following their settings, we fix the margin to 0.4 and tune the number of batches to pool for selecting negative examples from $\{40,60,80,100\}$. For both models, we set the dimension of the embeddings to 1024. For \lstmavg, we train a single layer bidirectional LSTM with hidden states of 512 dimensions. To create the sentence embedding, the forward and backward hidden states are concatenated and mean-pooled. Following~\citet{wieting-17-full}, we shuffle the inputs with probability $p$, tuning $p$ from $\{0.3, 0.5\}$.

We also implicitly compare to previous machine translation approaches like~\citet{espana2017empirical,schwenk2017learning,artetxe2018massively} in Appendix~\ref{appendix:embedding} where we explore different variations of training LSTM sequence-to-sequence models. We find that our translation baselines reported in the tables below (both LSTM and Transformer) outperform the architectures from these works due to using the Attention and Logit methods mentioned in Section~\ref{sec:models}%
, demonstrating that our baselines represent, or even over-represent, the state-of-the-art for machine translation approaches.

\paragraph{\vae Ablations}

Lastly, we compare to ablations of our model to better understand the benefits of parallel data, language-specific variables, the KL loss term, and how much we gain from the more conventional translation baselines.

\begin{itemize}[leftmargin=4mm]
\itemsep-0.5mm
    \item \monoae: English autoencoder on the English side of our \texttt{en-fr} data.
    \item \monovae: English variational autoencoder on the English side of our \texttt{en-fr} data.
    \item \engtrans: Translation from \texttt{en} to \texttt{fr}.
    \item \multitrans: Translation from both \texttt{en} to \texttt{fr} and \texttt{fr} to \texttt{en} where the encoding parameters are shared but each language has its own decoder. 
    \item \multitransprior: A model similar to \multitrans, but it includes a prior over the embedding space and therefore a KL loss term. This model differs from \vae since it does not have any language-specific variables.
    \item \trans: Follows the same architecture as \vae, but without the priors and KL loss term.
\end{itemize}

\subsection{Experimental Settings}

\begin{table*}
    \centering
    \small
    \begin{tabular}{| c | p{4.5cm} | p{4.5cm} | c | c |}
    \hline
    Data & Sentence 1 & Sentence 2 &  Gold Score \\
    \hline
    Hard+ & Other ways are needed. & It is necessary to find other means. & 4.5 \\
    Hard- & How long can you keep chocolate in the freezer? & How long can I keep bread dough in the refrigerator? & 1.0 \\
    Negation & It's not a good idea. & It's a good idea to do both. & 1.0\\
    \hline
    \end{tabular}
    \caption{Examples from our {\it Hard STS} dataset and our negation split. The sentence pair in the first row has dissimilar structure and vocabulary yet a high gold score. The second sentence pair has similar structure and vocabulary and a low gold score. The last sentence pair contains negation, where there is a {\it not} in Sentence 1 that causes otherwise similar sentences to have low semantic similarity.
    }
    \label{table:examples}
\end{table*}

The training data for our models is a mixture of OpenSubtitles 2018%
\footnote{\url{http://opus.nlpl.eu/OpenSubtitles.php}} \texttt{en}-\texttt{fr} data and \texttt{en}-\texttt{fr} Gigaword\footnote{\url{https://www.statmt.org/wmt10/training-giga-fren.tar}} data. To create our dataset, we combined the complete corpora of each dataset and then randomly selected 1,000,000 sentence pairs to be used for training with 10,000 used for validation. We use \texttt{sentencepiece}~\citep{kudo2018sentencepiece} with a vocabulary size of 20,000 to segment the sentences, and we chose sentence pairs whose sentences are between 5 and 100 tokens each.

In designing the model architectures for the encoders and decoders, we experimented with Transformers and LSTMs. Due to better performance, we use a 5 layer Transformer for each of the encoders and a single layer decoder for each of the decoders. This design decision was empirically motivated as we found using a larger decoder was slower and worsened performance, but conversely, adding more encoder layers improved performance. More discussion of these trade-offs along with ablations and comparisons to LSTMs are included in Appendix~\ref{appendix:ablations}.

For all of our models, we set the dimension of the embeddings and hidden states for the encoders and decoders to 1024. Since we experiment with two different architectures,\footnote{We use LSTMs in our ablations.} we follow two different optimization strategies. For training models with Transformers, we use Adam~\citep{kingma2014adam} with $\beta_1=0.9$, $\beta_2=0.98$, and $\epsilon=10^{-8}$. We use the same learning rate schedule as~\citet{vaswani2017attention}, i.e., the learning rate increases linearly for 4,000 steps to $5\times10^{-4}$, after which it is decayed proportionally to the inverse square root of the number of steps. For training the LSTM models, we use Adam with a fixed learning rate of 0.001. We train our models for 20 epochs.

For models incorporating a translation loss, we used label smoothed cross entropy~\citep{szegedy2016rethinking,pereyra2017regularizing} with $\epsilon = 0.1$. For \monovae, \vae and \multitrans, we anneal the KL term so that it increased linearly for $2^{16}$ updates, which robustly gave good results in preliminary experiments. We also found that in training BGT, combining its loss with the \multitrans objective during training of both models increased performance, and so this loss was summed with the BGT loss in all of our experiments.
We note that this does not affect our claim of BGT being a generative model, as this loss is only used in a multi-task objective at training time, and we calculate the generation probabilities according to standard BGT at test time.

Lastly, in Appendix~\ref{appendix:batch}, we illustrate that it is crucial to train the Transformers with large batch sizes. Without this, the model can learn the goal task (such as translation) with reasonable accuracy, but the learned semantic embeddings are of poor quality until batch sizes approximately reach 25,000 tokens. Therefore, we use a maximum batch size of 50,000 tokens in our \engtrans, \multitrans, and \trans, experiments and 25,000 tokens in our \multitransprior and \vae experiments.

\begin{table*}
\small
\centering
\setlength{\tabcolsep}{3pt}
\begin{tabular} { | l | c | c | c | c | c | c || c | c | c |}
\hline
\multirow{2}{*}{Model}  & \multicolumn{9}{c|}{Semantic Textual Similarity (STS)}\\
 \cline{2-10}
& 2012 & 2013 & 2014 & 2015 & 2016 & \bf Avg. & Hard+ & Hard- & \bf Avg. \\
\hline
BERT (CLS) & 33.2 & 29.6 & 34.3 & 45.1 & 48.4 & 38.1 &  7.8 & 12.5 & 10.2\\
BERT (Mean) & 48.8 & 46.5 & 54.0 & 59.2 & 63.4 & 54.4 & 3.1 & 24.1 & 13.6\\
Infersent & 61.1 & 51.4 & 68.1 & 70.9 & 70.7 & 64.4 & 4.2 & 29.6 & 16.9\\
GenSen & 60.7 & 50.8 & 64.1 & 73.3 & 66.0 & 63.0 & \bf 24.2 & 6.3 & 15.3\\
USE & 61.4 & 59.0 & 70.6 & 74.3 & 73.9 & 67.8 & 16.4 & 28.1 & 22.3\\
LASER & 63.1 & 47.0 & 67.7 & 74.9 & 71.9 & 64.9 & 18.1 & 23.8 & 20.9 \\
Sentence-BERT & 66.9 & \bf 63.2 & 74.2 & 77.3 & 72.8 & 70.9 & 23.9 & 3.6 & 13.8\\
\hline
\spavg & 68.4 & 60.3 & 75.1 & 78.7 & 76.8 & 71.9 & 19.1 & 29.8 & 24.5\\
\lstmavg & 67.9 & 56.4 & 74.5 & 78.2 & 75.9 & 70.6 & 18.5 & 23.2 & 20.9\\
\hline
\monoae & 60.2 & 52.7 & 68.6 & 74.0 & 73.2 & 65.7 & 15.7 & 36.0 & 25.9\\
\monovae & 59.5 & 54.0 & 67.3 & 74.6 & 74.1 & 65.9 & 16.8 & 42.7 & 29.8\\
\hline
\engtrans & 66.5 & 60.7 & 72.9 & 78.1 & 78.3 & 71.3 & 18.0 & 47.2 & 32.6\\
\multitrans & 67.1 & 61.0 & 73.3 & 78.0 & 77.8 & 71.4 & 20.0 & \bf 48.2 & 34.1\\
\multitransprior & 68.3 & 61.3 & 74.5 & 79.0 & 78.5 & 72.3 & 24.1 & 46.8 & \bf 35.5\\
\trans & 67.6 & 59.8 & 74.1 & 78.4 & 77.9 & 71.6 & 17.9 & 45.5 & 31.7\\
\vae & \bf 68.9 & 62.2 & \bf 75.9 & \bf 79.4 & \bf 79.3 & \bf 73.1 & 22.5 & 46.6 & 34.6\\
\hline
\end{tabular}
\caption{\label{table:results} Results of our models and models from prior work. The first six rows are pretrained models from the literature, the next two rows are strong baselines trained on the same data as our models, and the last seven rows include model ablations and \vae, our final model. 
We show results, measured in Pearson's $r \times 100$, for each year of the STS tasks 2012-2016 and our two {\it Hard STS} datasets.
}
\end{table*}

\begin{table*}
\centering
\small
\begin{tabular} { | l | c | c || c | c | c |}
\hline
Model & \texttt{es-es} & \texttt{ar-ar} & \texttt{en-es} & \texttt{en-ar} & \texttt{en-tr} \\
\hline
LASER & 79.7 & 69.3 & 59.7 & 65.5 & 72.0 \\
\hline
\multitrans & 83.4 & 72.6 & 64.1 & 37.6 & 59.1\\
\multitransprior & 81.7 & 72.8 & 72.6 & 73.4 & 74.8\\
\trans & 84.5 & 73.2 & 68.0 & 66.5 & 70.9\\
\vae & \bf 85.7 & \bf 74.9 & \bf 75.6 & \bf 73.5 & \bf 74.9\\
\hline
\end{tabular}
\caption{\label{table:crosslingual} Performance measured in Pearson's $r \times 100$, on the SemEval 2017 STS task on the \texttt{es-es}, \texttt{ar-ar}, \texttt{en-es}, \texttt{en-ar}, and \texttt{en-tr} datasets.
}
\end{table*}

\subsection{Evaluation} \label{sec:evaluation}

Our primary evaluation are the 2012-2016 SemEval Semantic Textual Similarity (STS) shared tasks~\citep{agirre2012semeval,agirre2013sem,agirre2014semeval,agirre2015semeval,agirre2016semeval}, where the goal is to accurately predict the degree to which two sentences have the same meaning as measured by human judges. The evaluation metric is Pearson's $r$ with the gold labels.

Secondly, we evaluate on {\it Hard STS}, where we combine and filter the STS datasets in order to make a more difficult evaluation. We hypothesize that these datasets contain many examples where their gold scores are easy to predict by either having similar structure and word choice and a high score or dissimilar structure and word choice and a low score. Therefore, we split the data using symmetric word error rate (SWER),\footnote{We define symmetric word error rate for sentences $s_1$ and $s_2$ as $\frac{1}{2}WER(s_1, s_2)$ + $\frac{1}{2}WER(s_2, s_1)$, since word error rate (WER) is an asymmetric measure.} finding sentence pairs with low SWER and low gold scores as well as sentence pairs with high SWER and high gold scores.  This results in two datasets, Hard+ which have SWERs in the bottom 20\% of all STS pairs and whose gold label is between 0 and 1,\footnote{STS scores are between 0 and 5.} and Hard- where the SWERs are in the top 20\% of the gold scores are between 4 and 5.
We also evaluate on a split where negation was likely present in the example.\footnote{We selected examples for the negation split where one sentence contained {\it not} or {\it 't} and the other did not.}
Examples are shown in Table~\ref{table:examples}.

Lastly, we evaluate on STS in \texttt{es} and \texttt{ar} as well as cross-lingual evaluations for \texttt{en-es}, \texttt{en-ar}, and \texttt{en-tr}. We use the datasets from SemEval 2017~\citep{cer2017semeval}. For this setting, we train \multitrans and \vae on 1 million examples from \texttt{en-es}, \texttt{en-ar}, and \texttt{en-tr} OpenSubtitles 2018 data.

\subsection{Results}

\begin{table*}
\centering
\footnotesize
\begin{tabular} { | l | c | c | c | c | c | c | c |}
\hline
\multirow{2}{*}{Model} & \multicolumn{7}{c|}{Semantic Textual Similarity (STS)}\\
\cline{2-8}
& 2012 & 2013 & 2014 & 2015 & 2016 & Hard+ & Hard- \\
\hline
Random Encoder & 51.4 & 34.6 & 52.7 & 52.3 & 49.7 & 4.8 & 17.9 \\
English Language Encoder & 44.4 & 41.7 & 53.8 & 62.4 & 61.7 & 15.3 & 26.5 \\
Semantic Encoder & \bf 68.9 & \bf 62.2 & \bf 75.9 & \bf 79.4 & \bf 79.3 & \bf 22.5 & \bf 46.6 \\
\hline
\end{tabular}
\caption{\label{table:garbage} STS performance on the 2012-2016 datasets and our {\it STS Hard} datasets for a randomly initialized Transformer, the trained English language-specific encoder from \vae, and the trained semantic encoder from \vae. Performance is measured in Pearson's $r \times 100$.
}
\end{table*}

\begin{table*}
\centering
\footnotesize
\setlength{\tabcolsep}{2pt}
\begin{tabular} { | l | l | c | c | c | c | c | c | c | c | c | c |}
\hline
Lang. & Model & STS & S. Num. & O. Num. & Depth & Top Con. & Word & Len. & P. Num. & P. First & Gend.\\
\hline
\multirow{4}{*}{\texttt{fr}}
& \multitrans & 71.2 & 78.0 & 76.5 & 28.2 & 65.9 & \bf 80.2 & 74.0 & 56.9 & 88.3 & 53.0\\
& Semantic Encoder & \bf 72.4 & \bf 84.6 & \bf 80.9 & \bf 29.7 & \bf 70.5 & 77.4 & 73.0 & 60.7 & 87.9 &  52.6\\
& \texttt{en} Language Encoder & 56.8 & 75.2 & 72.0 & 28.0 & 63.6 & 65.4 & \bf 80.2 & \bf 65.3 & \bf 92.2 & -\\
& \texttt{fr} Language Encoder &  -  &  -  &  -  &  -  &  -  &  -  &  -  &  -  &  -  &  \bf 56.5\\
\hline
\multirow{4}{*}{\texttt{es}} & \multitrans  & 70.5 & 84.5 & 82.1 & 29.7 & 68.5 &  \bf 79.2  & 77.7 & 63.4 & 90.1 &  54.3\\
& Semantic Encoder & \bf 72.1 & \bf 85.7 & \bf 83.6 & \bf 32.5 & \bf 71.0 & 77.3 & 76.7 & 63.1 & 89.9 &  52.6\\
& \texttt{en} Language Encoder & 55.8 & 75.7 & 73.7 & 29.1 & 63.9 & 63.3 & \bf 80.2 & \bf 64.2 & \bf 92.7  &  -\\
& \texttt{es} Language Encoder &  -  &  -  &  -  &  -  &  -  &  -  &  -  &  -  &  -  &  \bf 54.7\\
\hline
\multirow{3}{*}{\texttt{ar}}  
& \multitrans & 70.2 & 77.6 & 74.5 & 28.1 & 67.0 & \bf 77.5 & 72.3 & 57.5 & 89.0 & -\\
& Semantic Encoder & \bf 70.8 & \bf 81.9 & \bf 80.8 & \bf 32.1 & \bf 71.7 & 71.9 & 73.3 & 61.8 & 88.5 &  -\\
& \texttt{en} Language Encoder & 58.9 & 76.2 & 73.1 & 28.4 & 60.7 & 71.2 & \bf 79.8 & \bf 63.4 & \bf 92.4  &  -\\
\hline
\multirow{3}{*}{\texttt{tr}}
& \multitrans & 70.7 & 78.5 & 74.9 & 28.1 & 60.2 & \bf 78.4 & 72.1 & 54.8 & 87.3 &  - \\
& Semantic Encoder & \bf 72.3 & \bf 81.7 & \bf 80.2 & \bf 30.6 & \bf 66.0 & 75.2 & 72.4 & 59.3 & 86.7 &  -\\
& \texttt{en} Language Encoder & 57.8 & 77.3 & 74.4 & 28.3 & 63.1 & 67.1 & \bf 79.7 & \bf 67.0 & \bf 92.5  &  -\\
\hline
\multirow{3}{*}{\texttt{ja}}
& \multitrans & 71.0 & 66.4 & 64.6 & 25.4 & 54.1 & \bf 76.0 & 67.6 & 53.8 & 87.8 &  -\\
& Semantic Encoder & \bf 71.9  & 68.0 & 66.8 & 27.5 & 58.9 & 70.1 & 68.7 & 52.9 & 86.6 &  -\\
& \texttt{en} Language Encoder & 60.6 &  \bf 77.6  &  \bf 76.4  &  \bf 28.0  &  \bf 64.6  & 70.0 &  \bf 80.4  &  \bf 62.8  &  \bf 92.0  &  -\\
\hline
\end{tabular}
\vspace{-3mm}
\caption{\label{table:probe} Average STS performance for the 2012-2016 datasets, measured in Pearson's $r \times 100$, followed by probing results on predicting number of subjects, number of objects, constituent tree depth, top constituent, word content, length, number of punctuation marks, the first punctuation mark, and whether the articles in the sentence are the correct gender. All probing results are measured in accuracy $\times 100$.
}
\vspace{-6mm}
\end{table*}

The results on the STS and {\it Hard STS} are shown in Table~\ref{table:results}.\footnote{We obtained values for STS 2012-2016 from prior works using SentEval~\citep{conneau2018senteval}. Note that we include all datasets for the 2013 competition, including SMT, which is not included in SentEval.}
From the results, we see that \vae has the highest overall performance. It does especially well compared to prior work on the two {\it Hard STS} datasets. We used paired bootstrap resampling to check whether \vae significantly outperforms SentenceBert, \spavg, \multitrans, and \trans on the STS task. We found all gains to be \emph{significant with $p < 0.01$}.
\footnote{
We show further difficult splits in Appendix~\ref{appendix:hardsts}, including a negation split, beyond those used in {\it Hard STS} and compare the top two performing models in the STS task from Table~\ref{table:results}. We also show easier splits of the data to illustrate that the difference between these models is smaller on these splits.
}

From these results, we see that both positive examples that have little shared vocabulary and structure and negative examples with significant shared vocabulary and structure benefit significantly from using a deeper architecture. Similarly, examples where negation occurs also benefit from our deeper model. These examples are difficult because more than just the identity of the words is needed to determine the relationship of the two sentences, and this is something that \spavg is not equipped for since it is unable to model word order. The bottom two rows show {\it easier} examples where positive examples have high overlap and low SWER and vice versa for negative examples. Both models perform similarly on this data, with the \vae model having a small edge consistent with the overall gap between these two models.

Lastly, in Table~\ref{table:crosslingual}, we show the results of STS evaluations in \texttt{es} and \texttt{ar} and cross-lingual evaluations for \texttt{en-es}, \texttt{en-ar}, and \texttt{en-tr}. We also include a comparison to LASER, which is a multilingual model.\footnote{We note that this is not a fair comparison for a variety of reasons. For instance, our models are just trained on two languages at a time, but are only trained on 1M translation pairs from OpenSubtitles. LASER, in turn, is trained on 223M translation pairs covering more domains, but is also trained on 93 languages simultaneously.} From these results, we see that \vae has the best performance across all datasets, however the performance is significantly stronger than the \multitrans and \trans baselines in the cross-lingual setting. Since \multitransprior also has significantly better performance on these tasks, most of this gain seems to be due to the prior having a regularizing effect. However, \vae outperforms \multitransprior overall, and we hypothesize that the gap in performance between these two models is due to \vae being able to strip away the language-specific information in the representations with its language-specific variables, allowing for the semantics of the sentences to be more directly compared.

\section{Analysis} \label{sec:analysis}
We next analyze our \vae model by examining what elements of syntax and semantics the language and semantic variables capture relative both to each-other and to the sentence embeddings from the \multitrans models. We also analyze how the choice of language and its lexical and syntactic distance from English affects the semantic and syntactic information captured by the semantic and language-specific encoders. Finally, we also show that our model is capable of sentence generation in a type of {\it style transfer}, demonstrating its capabilities as a generative model.

\subsection{STS}

We first show that the language variables are capturing little semantic information by evaluating the learned English language-specific variable from our \vae model on our suite of semantic tasks. The results in Table~\ref{table:garbage} show that these encoders perform closer to a random encoder than the semantic encoder from \vae. This is consistent with what we would expect to see if they are capturing extraneous language-specific information.

\subsection{Probing}

We probe our \vae semantic and language-specific encoders, along with our \multitrans encoders as a baseline, to compare and contrast what aspects of syntax and semantics they are learning relative to each other across five languages with various degrees of similarity with English. All models are trained on the OpenSubtitles 2018 corpus. We use the datasets from~\citet{conneau2018you} for semantic tasks like number of subjects and number of objects, and syntactic tasks like tree depth, and top constituent. Additionally, we include predicting the word content and sentence length. We also add our own tasks to validate our intuitions about punctuation and language-specific information. In the first of these, {\it punctuation number}, we train a classifier to predict the number of punctuation marks\footnote{Punctuation were taken from the set 
\{ ' ! " \# \$ \% \& \textbackslash 
' ( ) $*$ $+$ , $-$ . $/$ : ; $<$ 
$=$ $>$ ? @ [ ] \^ \_ ` \{ 
| \} \~ ' . \}.} in a sentence. To make the task more challenging, we limit each label to have at most 20,000 examples split among training, validation, and testing data.\footnote{The labels are from 1 punctuation mark up to 10 marks with an additional label consolidating 11 or more marks.} In the second task, {\it punctuation first}, we train a classifier to predict the identity of the first punctuation mark in the sentence. In our last task, {\it gender}, we detect examples where the gender of the articles in the sentence is incorrect in French or Spanish.
To create an incorrect example, we switch articles from $\{$le, la, un, une$\}$ for French and $\{$el, la, los, las$\}$ for Spanish, with their (indefinite or definite for French and singular or plural for Spanish) counterpart with the opposite gender. This dataset was balanced so random chance gives 50\% on the testing data.  All tasks use 100,000 examples for training and 10,000 examples for validation and testing. The results of these experiments are shown in Table~\ref{table:probe}.

These results show that the source separation is effective - stylistic and language-specific information like length, punctuation and language-specific gender information are more concentrated in the language variables, while word content, semantic and syntactic information are more concentrated in the semantic encoder. The choice of language is also seen to be influential on what these encoders are capturing. When the languages are closely related to English, like in French and Spanish, the performance difference between the semantic and English language encoder is larger for word content, subject number, object number than for more distantly related languages like Arabic and Turkish. In fact, word content performance is directly tied to how well the alphabets of the two languages overlap. This relationship matches our intuition, because lexical information will be cheaper to encode in the semantic variable when it is shared between the languages.
Similarly for the tasks of length, punctuation first, and punctuation number, the gap in performance between the two encoders also grows as the languages become more distant from English. Lastly, the gap on STS performance between the two encoders shrinks as the languages become more distant, which again is what we would expect, as the language-specific encoders are forced to capture more information. 

Japanese is an interesting case in these experiments, where the English language-specific encoder outperforms the semantic encoder on the semantic and syntactic probing tasks. Japanese is a very distant language to English both in its writing system and in its sentence structure (it is an SOV language, where English is an SVO language). However, despite these difference, the semantic encoder strongly outperforms the English language-specific encoder, suggesting that the underlying meaning of the sentence is much better captured by the semantic encoder.

\subsection{Generation and Style Transfer}

\begin{table}[h!]
    \centering
    \footnotesize
    \begin{tabular}{| l | p{0.75\columnwidth} |}
    \hline
    Source & you know what i've seen? \\
    Style & he said, ``since when is going fishing" had anything to do with fish?'' \\
    Output & he said, ``what is going to do with me since i saw you?'' \\
    \hline
    Source & guys, that was the tech unit. \\
    Style & is well, ``capicci'' ... \\
    Output & is that what, ``technician''? \\
    \hline
    Source & the pay is no good, but it's money. \\
    Style & do we know cause of death? \\
    Output & do we have any money? \\
    \hline
    Source & we're always doing stupid things. \\
    Style & all right listen, i like being exactly where i am, \\
    Output & all right, i like being stupid, but i am always here. \\
    \hline
    \end{tabular}
    \caption{{\it Style transfer} generations from our learned \vae model. {\it Source} refers to the sentence fed into the semantic encoder, {\it Style} refers to the sentence fed into the English language-specific encoder, and {\it Output} refers to the text generated by our model.
    }
    \label{table:style}
    \vspace{-6mm}
\end{table}

In this section, we qualitatively demonstrate the ability of our model to generate sentences. We focus on a {\it style-transfer} task where we have original seed sentences from which we calculate our semantic vector $z_{sem}$ and language specific vector $z_{en}$.
Specifically, we feed in a {\it Source} sentence into the semantic encoder to obtain $z_{sem}$, and another {\it Style} sentence into the English language-specific encoder to obtain $z_{en}$. We then generate a new sentence using these two latent variables.
This can be seen as a type of style transfer where we expect the model to generate a sentence that has the semantics of the {\it Source} sentence and the style of the {\it Style} sentence. We use our \texttt{en-fr} \vae model from Table~\ref{table:probe} and show some examples in Table~\ref{table:style}. All input sentences are from held-out \texttt{en-fr} OpenSubtitles data. From these examples, we see further evidence of the role of the semantic and language-specific encoders, where most of the semantics (e.g. topical word such as {\it seen} and {\it tech} in the {\it Source} sentence) are reflected in the output, but length and structure are more strongly influenced by the language-specific encoder.

\section{Conclusion}
We propose Bilingual Generative Transformers, a model that uses parallel data to learn to perform {\it source separation} of common semantic information between two languages from language-specific information. We show that the model is able to accomplish this source separation through probing tasks and text generation in a style-transfer setting. We find that our model bests all baselines on unsupervised semantic similarity tasks, with the largest gains coming from a new challenge we propose as {\it Hard STS}, designed to foil methods approximating semantic similarity as word overlap. We also find our model to be especially effective on unsupervised cross-lingual semantic similarity, due to its stripping away of language-specific information allowing for the underlying semantics to be more directly compared. In future work, we will explore generalizing this approach to the multilingual setting, or applying it to the pre-train and fine-tune paradigm used widely in other models such as BERT.

\section*{Acknowledgments}

The authors thank Amazon AWS for the generous donation of computation credits. This project is funded in part by the NSF under grants 1618044 and 1936155, and by the NEH under grant HAA256044-17.

\bibliographystyle{acl_natbib}
\bibliography{emnlp2020}

\begin{thebibliography}{50}
\expandafter\ifx\csname natexlab\endcsname\relax\def\natexlab#1{#1}\fi

\bibitem[{Agirre et~al.(2015)Agirre, Banea, Cardie, Cer, Diab, Gonzalez-Agirre,
  Guo, Lopez-Gazpio, Maritxalar, Mihalcea, Rigau, Uria, and
  Wiebe}]{agirre2015semeval}
Eneko Agirre, Carmen Banea, Claire Cardie, Daniel Cer, Mona Diab, Aitor
  Gonzalez-Agirre, Weiwei Guo, Inigo Lopez-Gazpio, Montse Maritxalar, Rada
  Mihalcea, German Rigau, Larraitz Uria, and Janyce Wiebe. 2015.
\newblock {SemEval}-2015 task 2: Semantic textual similarity, {English},
  {Spanish} and pilot on interpretability.
\newblock In \emph{Proceedings of the 9th International Workshop on Semantic
  Evaluation ({SemEval} 2015)}.

\bibitem[{Agirre et~al.(2014)Agirre, Banea, Cardie, Cer, Diab, Gonzalez-Agirre,
  Guo, Mihalcea, Rigau, and Wiebe}]{agirre2014semeval}
Eneko Agirre, Carmen Banea, Claire Cardie, Daniel Cer, Mona Diab, Aitor
  Gonzalez-Agirre, Weiwei Guo, Rada Mihalcea, German Rigau, and Janyce Wiebe.
  2014.
\newblock {SemEval}-2014 task 10: Multilingual semantic textual similarity.
\newblock In \emph{Proceedings of the 8th International Workshop on Semantic
  Evaluation ({SemEval} 2014)}.

\bibitem[{Agirre et~al.(2016)Agirre, Banea, Cer, Diab, Gonzalez-Agirre,
  Mihalcea, Rigau, and Wiebe}]{agirre2016semeval}
Eneko Agirre, Carmen Banea, Daniel Cer, Mona Diab, Aitor Gonzalez-Agirre, Rada
  Mihalcea, German Rigau, and Janyce Wiebe. 2016.
\newblock {SemEval}-2016 task 1: Semantic textual similarity, monolingual and
  cross-lingual evaluation.
\newblock \emph{Proceedings of SemEval}, pages 497--511.

\bibitem[{Agirre et~al.(2013)Agirre, Cer, Diab, Gonzalez-Agirre, and
  Guo}]{agirre2013sem}
Eneko Agirre, Daniel Cer, Mona Diab, Aitor Gonzalez-Agirre, and Weiwei Guo.
  2013.
\newblock * sem 2013 shared task: Semantic textual similarity.
\newblock In \emph{Second Joint Conference on Lexical and Computational
  Semantics (* SEM), Volume 1: Proceedings of the Main Conference and the
  Shared Task: Semantic Textual Similarity}, volume~1, pages 32--43.

\bibitem[{Agirre et~al.(2012)Agirre, Diab, Cer, and
  Gonzalez-Agirre}]{agirre2012semeval}
Eneko Agirre, Mona Diab, Daniel Cer, and Aitor Gonzalez-Agirre. 2012.
\newblock {SemEval}-2012 task 6: A pilot on semantic textual similarity.
\newblock In \emph{Proceedings of the First Joint Conference on Lexical and
  Computational Semantics-Volume 1: Proceedings of the main conference and the
  shared task, and Volume 2: Proceedings of the Sixth International Workshop on
  Semantic Evaluation}. Association for Computational Linguistics.

\bibitem[{Arora et~al.(2017)Arora, Liang, and Ma}]{arora2017simple}
Sanjeev Arora, Yingyu Liang, and Tengyu Ma. 2017.
\newblock A simple but tough-to-beat baseline for sentence embeddings.
\newblock In \emph{Proceedings of the International Conference on Learning
  Representations}.

\bibitem[{Artetxe and Schwenk(2018)}]{artetxe2018massively}
Mikel Artetxe and Holger Schwenk. 2018.
\newblock Massively multilingual sentence embeddings for zero-shot
  cross-lingual transfer and beyond.
\newblock \emph{arXiv preprint arXiv:1812.10464}.

\bibitem[{Bowman et~al.(2015)Bowman, Angeli, Potts, and
  Manning}]{bowman2015large}
Samuel~R. Bowman, Gabor Angeli, Christopher Potts, and Christopher~D. Manning.
  2015.
\newblock A large annotated corpus for learning natural language inference.
\newblock In \emph{Proceedings of the 2015 Conference on Empirical Methods in
  Natural Language Processing}, pages 632--642, Lisbon, Portugal.

\bibitem[{Cer et~al.(2017)Cer, Diab, Agirre, Lopez-Gazpio, and
  Specia}]{cer2017semeval}
Daniel Cer, Mona Diab, Eneko Agirre, Inigo Lopez-Gazpio, and Lucia Specia.
  2017.
\newblock {SemEval-2017} {T}ask 1: Semantic textual similarity multilingual and
  crosslingual focused evaluation.
\newblock In \emph{Proceedings of the 11th International Workshop on Semantic
  Evaluation (SemEval-2017)}, pages 1--14, Vancouver, Canada.

\bibitem[{Cer et~al.(2018)Cer, Yang, Kong, Hua, Limtiaco, John, Constant,
  Guajardo-Cespedes, Yuan, Tar et~al.}]{cer2018universal}
Daniel Cer, Yinfei Yang, Sheng-yi Kong, Nan Hua, Nicole Limtiaco, Rhomni~St
  John, Noah Constant, Mario Guajardo-Cespedes, Steve Yuan, Chris Tar, et~al.
  2018.
\newblock Universal sentence encoder.
\newblock \emph{arXiv preprint arXiv:1803.11175}.

\bibitem[{Chen et~al.(2019)Chen, Tang, Wiseman, and Gimpel}]{chen2019multi}
Mingda Chen, Qingming Tang, Sam Wiseman, and Kevin Gimpel. 2019.
\newblock A multi-task approach for disentangling syntax and semantics in
  sentence representations.
\newblock \emph{arXiv preprint arXiv:1904.01173}.

\bibitem[{Chidambaram et~al.(2019)Chidambaram, Yang, Cer, Yuan, Sung, Strope,
  and Kurzweil}]{chidambaram-etal-2019-learning}
Muthu Chidambaram, Yinfei Yang, Daniel Cer, Steve Yuan, Yunhsuan Sung, Brian
  Strope, and Ray Kurzweil. 2019.
\newblock \href {https://doi.org/10.18653/v1/W19-4330} {Learning cross-lingual
  sentence representations via a multi-task dual-encoder model}.
\newblock In \emph{Proceedings of the 4th Workshop on Representation Learning
  for NLP (RepL4NLP-2019)}, pages 250--259, Florence, Italy. Association for
  Computational Linguistics.

\bibitem[{Conneau et~al.(2019)Conneau, Khandelwal, Goyal, Chaudhary, Wenzek,
  Guzm{\'a}n, Grave, Ott, Zettlemoyer, and Stoyanov}]{conneau2019unsupervised}
Alexis Conneau, Kartikay Khandelwal, Naman Goyal, Vishrav Chaudhary, Guillaume
  Wenzek, Francisco Guzm{\'a}n, Edouard Grave, Myle Ott, Luke Zettlemoyer, and
  Veselin Stoyanov. 2019.
\newblock Unsupervised cross-lingual representation learning at scale.
\newblock \emph{arXiv preprint arXiv:1911.02116}.

\bibitem[{Conneau and Kiela(2018)}]{conneau2018senteval}
Alexis Conneau and Douwe Kiela. 2018.
\newblock Senteval: An evaluation toolkit for universal sentence
  representations.
\newblock \emph{arXiv preprint arXiv:1803.05449}.

\bibitem[{Conneau et~al.(2017)Conneau, Kiela, Schwenk, Barrault, and
  Bordes}]{conneau2017supervised}
Alexis Conneau, Douwe Kiela, Holger Schwenk, Lo\"{i}c Barrault, and Antoine
  Bordes. 2017.
\newblock Supervised learning of universal sentence representations from
  natural language inference data.
\newblock In \emph{Proceedings of the 2017 Conference on Empirical Methods in
  Natural Language Processing}, pages 670--680, Copenhagen, Denmark.

\bibitem[{Conneau et~al.(2018)Conneau, Kruszewski, Lample, Barrault, and
  Baroni}]{conneau2018you}
Alexis Conneau, German Kruszewski, Guillaume Lample, Lo{\"\i}c Barrault, and
  Marco Baroni. 2018.
\newblock What you can cram into a single vector: Probing sentence embeddings
  for linguistic properties.
\newblock \emph{arXiv preprint arXiv:1805.01070}.

\bibitem[{Conneau and Lample(2019)}]{conneau2019cross}
Alexis Conneau and Guillaume Lample. 2019.
\newblock Cross-lingual language model pretraining.
\newblock In \emph{Advances in Neural Information Processing Systems}, pages
  7059--7069.

\bibitem[{Devlin et~al.(2018)Devlin, Chang, Lee, and
  Toutanova}]{devlin2018bert}
Jacob Devlin, Ming-Wei Chang, Kenton Lee, and Kristina Toutanova. 2018.
\newblock Bert: Pre-training of deep bidirectional transformers for language
  understanding.
\newblock \emph{arXiv preprint arXiv:1810.04805}.

\bibitem[{Dolan et~al.(2004)Dolan, Quirk, and Brockett}]{dolan-04}
Bill Dolan, Chris Quirk, and Chris Brockett. 2004.
\newblock Unsupervised construction of large paraphrase corpora: Exploiting
  massively parallel news sources.
\newblock In \emph{Proceedings of {COLING}}.

\bibitem[{Espana-Bonet et~al.(2017)Espana-Bonet, Varga, Barr{\'o}n-Cede{\~n}o,
  and van Genabith}]{espana2017empirical}
Cristina Espana-Bonet, Ad{\'a}m~Csaba Varga, Alberto Barr{\'o}n-Cede{\~n}o, and
  Josef van Genabith. 2017.
\newblock An empirical analysis of nmt-derived interlingual embeddings and
  their use in parallel sentence identification.
\newblock \emph{IEEE Journal of Selected Topics in Signal Processing},
  11(8):1340--1350.

\bibitem[{Ganitkevitch et~al.(2013)Ganitkevitch, Durme, and
  Callison-Burch}]{GanitkevitchDC13-short}
Juri Ganitkevitch, Benjamin~Van Durme, and Chris Callison-Burch. 2013.
\newblock {PPDB}: The {P}araphrase {D}atabase.
\newblock In \emph{Proceedings of HLT-NAACL}.

\bibitem[{He et~al.(2019)He, Spokoyny, Neubig, and
  Berg-Kirkpatrick}]{he2019lagging}
Junxian He, Daniel Spokoyny, Graham Neubig, and Taylor Berg-Kirkpatrick. 2019.
\newblock Lagging inference networks and posterior collapse in variational
  autoencoders.
\newblock \emph{arXiv preprint arXiv:1901.05534}.

\bibitem[{Hill et~al.(2016)Hill, Cho, and Korhonen}]{hill2016learning}
Felix Hill, Kyunghyun Cho, and Anna Korhonen. 2016.
\newblock Learning distributed representations of sentences from unlabelled
  data.
\newblock In \emph{Proceedings of the 2016 Conference of the North American
  Chapter of the Association for Computational Linguistics: Human Language
  Technologies}.

\bibitem[{Hochreiter and Schmidhuber(1997)}]{hochreiter1997long}
Sepp Hochreiter and J{\"u}rgen Schmidhuber. 1997.
\newblock Long short-term memory.
\newblock \emph{Neural computation}, 9(8).

\bibitem[{Kingma and Ba(2014)}]{kingma2014adam}
Diederik Kingma and Jimmy Ba. 2014.
\newblock Adam: A method for stochastic optimization.
\newblock \emph{arXiv preprint arXiv:1412.6980}.

\bibitem[{Kingma and Welling(2013)}]{kingma2013auto}
Diederik~P Kingma and Max Welling. 2013.
\newblock Auto-encoding variational bayes.
\newblock \emph{arXiv preprint arXiv:1312.6114}.

\bibitem[{Kiros et~al.(2015)Kiros, Zhu, Salakhutdinov, Zemel, Urtasun,
  Torralba, and Fidler}]{kiros2015skip}
Ryan Kiros, Yukun Zhu, Ruslan~R Salakhutdinov, Richard Zemel, Raquel Urtasun,
  Antonio Torralba, and Sanja Fidler. 2015.
\newblock Skip-thought vectors.
\newblock In \emph{Advances in Neural Information Processing Systems 28}, pages
  3294--3302.

\bibitem[{Kudo and Richardson(2018)}]{kudo2018sentencepiece}
Taku Kudo and John Richardson. 2018.
\newblock Sentencepiece: A simple and language independent subword tokenizer
  and detokenizer for neural text processing.
\newblock \emph{arXiv preprint arXiv:1808.06226}.

\bibitem[{Ma et~al.(2019)Ma, Zhou, Li, Neubig, and Hovy}]{ma2019flowseq}
Xuezhe Ma, Chunting Zhou, Xian Li, Graham Neubig, and Eduard Hovy. 2019.
\newblock \href {https://doi.org/10.18653/v1/D19-1437} {{F}low{S}eq:
  Non-autoregressive conditional sequence generation with generative flow}.
\newblock In \emph{Proceedings of the 2019 Conference on Empirical Methods in
  Natural Language Processing and the 9th International Joint Conference on
  Natural Language Processing (EMNLP-IJCNLP)}, pages 4273--4283, Hong Kong,
  China. Association for Computational Linguistics.

\bibitem[{Mikolov et~al.(2013)Mikolov, Sutskever, Chen, Corrado, and
  Dean}]{mikolov2013distributed}
Tomas Mikolov, Ilya Sutskever, Kai Chen, Greg~S. Corrado, and Jeff Dean. 2013.
\newblock Distributed representations of words and phrases and their
  compositionality.
\newblock In \emph{Advances in Neural Information Processing Systems}.

\bibitem[{Pennington et~al.(2014)Pennington, Socher, and
  Manning}]{pennington2014glove}
Jeffrey Pennington, Richard Socher, and Christopher~D. Manning. 2014.
\newblock Glove: Global vectors for word representation.
\newblock \emph{Proceedings of Empirical Methods in Natural Language Processing
  (EMNLP 2014)}.

\bibitem[{Pereyra et~al.(2017)Pereyra, Tucker, Chorowski, Kaiser, and
  Hinton}]{pereyra2017regularizing}
Gabriel Pereyra, George Tucker, Jan Chorowski, {\L}ukasz Kaiser, and Geoffrey
  Hinton. 2017.
\newblock Regularizing neural networks by penalizing confident output
  distributions.
\newblock \emph{arXiv preprint arXiv:1701.06548}.

\bibitem[{Peters et~al.(2018)Peters, Neumann, Iyyer, Gardner, Clark, Lee, and
  Zettlemoyer}]{peters2018deep}
Matthew~E Peters, Mark Neumann, Mohit Iyyer, Matt Gardner, Christopher Clark,
  Kenton Lee, and Luke Zettlemoyer. 2018.
\newblock Deep contextualized word representations.
\newblock In \emph{Proceedings of NAACL-HLT}, pages 2227--2237.

\bibitem[{Popel and Bojar(2018)}]{popel2018training}
Martin Popel and Ond{\v{r}}ej Bojar. 2018.
\newblock Training tips for the transformer model.
\newblock \emph{The Prague Bulletin of Mathematical Linguistics},
  110(1):43--70.

\bibitem[{Reimers and Gurevych(2019)}]{reimers2019sentence}
Nils Reimers and Iryna Gurevych. 2019.
\newblock Sentence-bert: Sentence embeddings using siamese bert-networks.
\newblock \emph{arXiv preprint arXiv:1908.10084}.

\bibitem[{Schwenk(2018)}]{schwenk2018filtering}
Holger Schwenk. 2018.
\newblock Filtering and mining parallel data in a joint multilingual space.
\newblock \emph{arXiv preprint arXiv:1805.09822}.

\bibitem[{Schwenk and Douze(2017)}]{schwenk2017learning}
Holger Schwenk and Matthijs Douze. 2017.
\newblock Learning joint multilingual sentence representations with neural
  machine translation.
\newblock \emph{arXiv preprint arXiv:1704.04154}.

\bibitem[{Subramanian et~al.(2018)Subramanian, Trischler, Bengio, and
  Pal}]{subramanian2018learning}
Sandeep Subramanian, Adam Trischler, Yoshua Bengio, and Christopher~J Pal.
  2018.
\newblock Learning general purpose distributed sentence representations via
  large scale multi-task learning.
\newblock \emph{arXiv preprint arXiv:1804.00079}.

\bibitem[{Szegedy et~al.(2016)Szegedy, Vanhoucke, Ioffe, Shlens, and
  Wojna}]{szegedy2016rethinking}
Christian Szegedy, Vincent Vanhoucke, Sergey Ioffe, Jon Shlens, and Zbigniew
  Wojna. 2016.
\newblock Rethinking the inception architecture for computer vision.
\newblock In \emph{Proceedings of the IEEE conference on computer vision and
  pattern recognition}, pages 2818--2826.

\bibitem[{Vaswani et~al.(2017)Vaswani, Shazeer, Parmar, Uszkoreit, Jones,
  Gomez, Kaiser, and Polosukhin}]{vaswani2017attention}
Ashish Vaswani, Noam Shazeer, Niki Parmar, Jakob Uszkoreit, Llion Jones,
  Aidan~N Gomez, {\L}ukasz Kaiser, and Illia Polosukhin. 2017.
\newblock Attention is all you need.
\newblock In \emph{Advances in Neural Information Processing Systems}, pages
  5998--6008.

\bibitem[{Wieting et~al.(2016{\natexlab{a}})Wieting, Bansal, Gimpel, and
  Livescu}]{wieting2016charagram}
John Wieting, Mohit Bansal, Kevin Gimpel, and Karen Livescu.
  2016{\natexlab{a}}.
\newblock Charagram: Embedding words and sentences via character $n$-grams.
\newblock In \emph{Proceedings of the 2016 Conference on Empirical Methods in
  Natural Language Processing}, pages 1504--1515.

\bibitem[{Wieting et~al.(2016{\natexlab{b}})Wieting, Bansal, Gimpel, and
  Livescu}]{wieting-16-full}
John Wieting, Mohit Bansal, Kevin Gimpel, and Karen Livescu.
  2016{\natexlab{b}}.
\newblock Towards universal paraphrastic sentence embeddings.
\newblock In \emph{Proceedings of the International Conference on Learning
  Representations}.

\bibitem[{Wieting et~al.(2019{\natexlab{a}})Wieting, Berg-Kirkpatrick, Gimpel,
  and Neubig}]{wieting2019beyond}
John Wieting, Taylor Berg-Kirkpatrick, Kevin Gimpel, and Graham Neubig.
  2019{\natexlab{a}}.
\newblock Beyond bleu: Training neural machine translation with semantic
  similarity.
\newblock In \emph{Proceedings of the 57th Annual Meeting of the Association
  for Computational Linguistics}, pages 4344--4355.

\bibitem[{Wieting and Gimpel(2017)}]{wieting-17-full}
John Wieting and Kevin Gimpel. 2017.
\newblock Revisiting recurrent networks for paraphrastic sentence embeddings.
\newblock In \emph{Proceedings of the 55th Annual Meeting of the Association
  for Computational Linguistics (Volume 1: Long Papers)}, pages 2078--2088,
  Vancouver, Canada.

\bibitem[{Wieting and Gimpel(2018)}]{wieting2017pushing}
John Wieting and Kevin Gimpel. 2018.
\newblock \href {http://aclweb.org/anthology/P18-1042} {{ParaNMT-50M}: Pushing
  the limits of paraphrastic sentence embeddings with millions of machine
  translations}.
\newblock In \emph{Proceedings of the 56th Annual Meeting of the Association
  for Computational Linguistics (Volume 1: Long Papers)}, pages 451--462.
  Association for Computational Linguistics.

\bibitem[{Wieting et~al.(2019{\natexlab{b}})Wieting, Gimpel, Neubig, and
  Berg-Kirkpatrick}]{wieting2019para}
John Wieting, Kevin Gimpel, Graham Neubig, and Taylor Berg-Kirkpatrick.
  2019{\natexlab{b}}.
\newblock Simple and effective paraphrastic similarity from parallel
  translations.
\newblock \emph{Proceedings of the ACL}.

\bibitem[{Yang et~al.(2020)Yang, Cer, Ahmad, Guo, Law, Constant, Abrego, Yuan,
  Tar, Sung, Strope, and Kurzweil}]{yang-etal-2020-multilingual}
Yinfei Yang, Daniel Cer, Amin Ahmad, Mandy Guo, Jax Law, Noah Constant,
  Gustavo~Hernandez Abrego, Steve Yuan, Chris Tar, Yun-hsuan Sung, Brian
  Strope, and Ray Kurzweil. 2020.
\newblock \href {https://doi.org/10.18653/v1/2020.acl-demos.12} {Multilingual
  universal sentence encoder for semantic retrieval}.
\newblock In \emph{Proceedings of the 58th Annual Meeting of the Association
  for Computational Linguistics: System Demonstrations}, pages 87--94, Online.
  Association for Computational Linguistics.

\bibitem[{Yang et~al.(2017)Yang, Hu, Salakhutdinov, and
  Berg-Kirkpatrick}]{yang2017improved}
Zichao Yang, Zhiting Hu, Ruslan Salakhutdinov, and Taylor Berg-Kirkpatrick.
  2017.
\newblock Improved variational autoencoders for text modeling using dilated
  convolutions.
\newblock In \emph{Proceedings of the 34th International Conference on Machine
  Learning-Volume 70}, pages 3881--3890. JMLR. org.

\bibitem[{Ziegler and Rush(2019)}]{ziegler2019latent}
Zachary Ziegler and Alexander Rush. 2019.
\newblock Latent normalizing flows for discrete sequences.
\newblock In \emph{International Conference on Machine Learning}, pages
  7673--7682.

\bibitem[{Zweigenbaum et~al.(2018)Zweigenbaum, Sharoff, and
  Rapp}]{zweigenbaum2018overview}
Pierre Zweigenbaum, Serge Sharoff, and Reinhard Rapp. 2018.
\newblock Overview of the third bucc shared task: Spotting parallel sentences
  in comparable corpora.
\newblock In \emph{Proceedings of 11th Workshop on Building and Using
  Comparable Corpora}, pages 39--42.

\end{thebibliography}

\newpage
-
\appendix

\section{Location of Sentence Embedding in Decoder for Learning Representations} \label{appendix:embedding}
As mentioned in Section~\ref{sec:models}, we experimented with 4 ways to incorporate the sentence embedding into the decoder: Word, Hidden, Attention, and Logit.  We also experimented with combinations of these 4 approaches. We evaluate these embeddings on the STS tasks and show the results, along with the time to train the models 1 epoch in Table~\ref{table:embedding}.

For these experiments, we train a single layer bidirectional LSTM (BiLSTM) \engtrans model with embedding size set to 1024 and hidden states set to 512 dimensions (in order to be roughly equivalent to our Transformer models). To form the sentence embedding in this variant, we mean pool the hidden states for each time step. The cell states of the decoder are initialized to the zero vector.

\begin{table}[h!]
\centering
\small
\begin{tabular} { | l | c | c |}
\hline
Architecture & STS & Time (s)\\
\hline
BiLSTM (Hidden) & 54.3 & 1226 \\
BiLSTM (Word) & 67.2 & 1341 \\
BiLSTM (Attention) & 68.8 & 1481 \\
BiLSTM (Logit) & 69.4 & 1603 \\
BiLSTM (Wd. + Hd.) & 67.3 & 1377 \\
BiLSTM (Wd. + Hd. + Att.) & 68.3 & 1669 \\
BiLSTM (Wd. + Hd. + Log.) & 69.1 & 1655 \\
BiLSTM (Wd. + Hd. + Att. + Log.) & 68.9 & 1856 \\
\hline
\end{tabular}
\caption{\label{table:embedding} Results for different ways of incorporating the sentence embedding in the decoder for a BiLSTM on the Semantic Textual Similarity (STS) datasets, along with the time taken to train the model for 1 epoch. Performance is measured in Pearson's $r \times 100$.
}
\end{table}

From this analysis, we see that the best performance is achieved with Logit, when the sentence embedding is place just prior to the softmax. The performance is much better than Hidden or Hidden+Word used in prior work. For instance, recently~\citep{artetxe2018massively} used the Hidden+Word strategy in learning multilingual sentence embeddings.

\subsection{VAE Training}

We also found that incorporating the latent code of a VAE into the decoder using the Logit strategy increases the mutual information while having little effect on the log likelihood. We trained two LSTM VAE models following the settings and aggressive training strategy in~\citep{he2019lagging}, where one LSTM model used the Hidden strategy and the other used the Hidden + Logit strategy. We trained the models on the \texttt{en} side of our \texttt{en-fr} data. We found that the mutual information increased form 0.89 to 2.46, while the approximate negative log likelihood, estimated by importance weighting, increased slightly from 53.3 to 54.0 when using Logit.

\section{Relationship Between Batch Size and Performance for Transformer and LSTM} \label{appendix:batch}
\begin{figure}[h]
    \centering
    \includegraphics[width=\columnwidth]{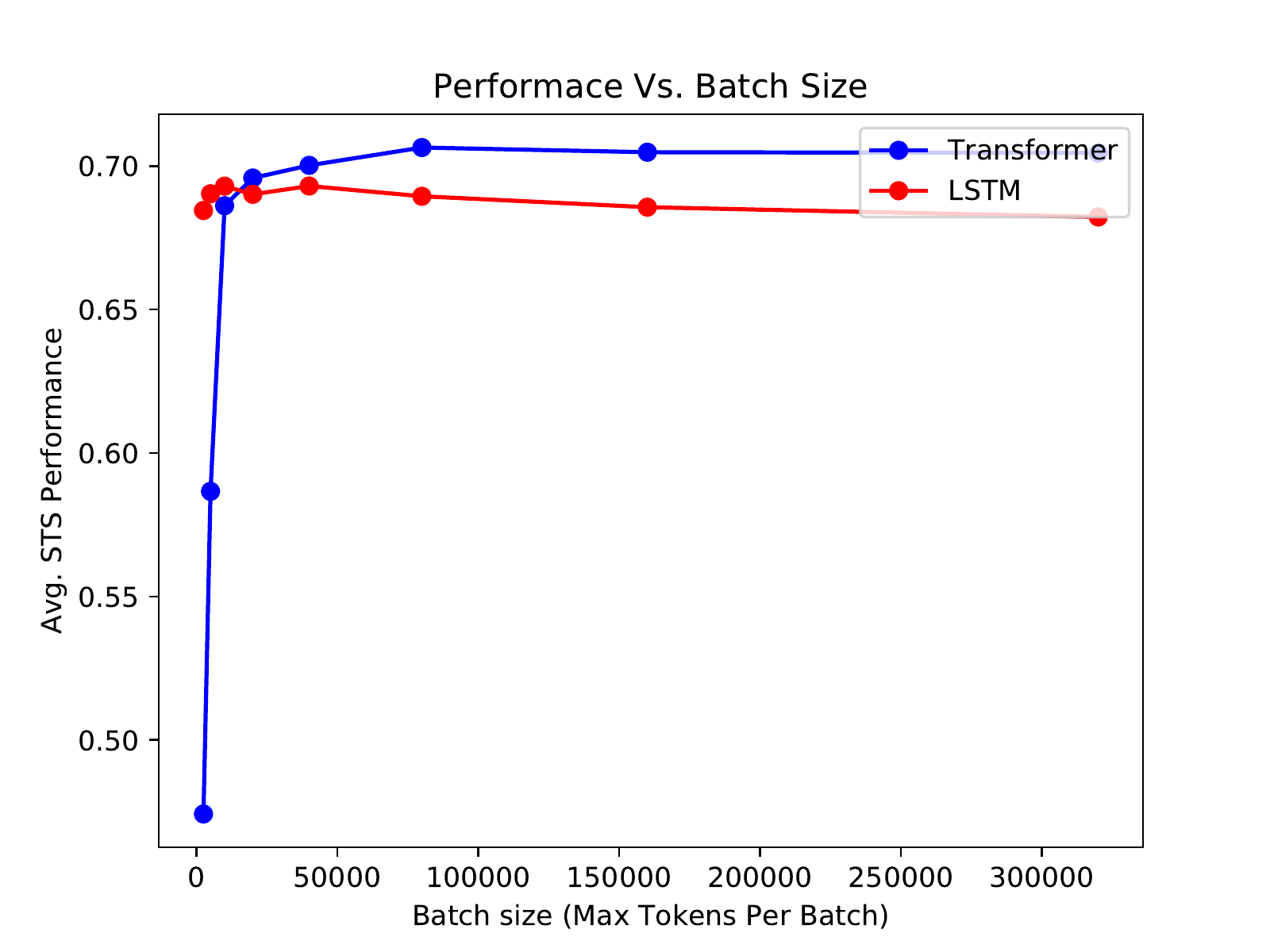}
    \caption{The relationship between average performance for each year of the STS tasks 2012-2016 (Pearson's $r \times 100$) and batch size (maximum number of words per batch).}
    \label{fig:batchsize}
\end{figure}

\begin{table*}[th!]
\centering
\begin{tabular} { | l | c | c | c |}
\hline
Data Split & $n$ & \vae & \spavg \\
\hline
All & 13,023 & \bf 75.3 & 74.1 \\
\hline
Negation & 705 & \bf 73.1 & 68.7 \\
\hline
Bottom 20\% SWER, label $ \in [0,2]$ & 404 & \bf 63.6 & 54.9 \\
Bottom 10\% SWER, label $ \in [0,1]$ & 72 & \bf 47.1 & 22.5 \\
Top 20\% SWER, label $ \in [3,5]$ & 937 & \bf 20.0 & 14.4 \\
Top 10\% SWER, label $ \in [4,5]$ & 159 & \bf 18.1 & 10.8 \\
\hline
Top 20\% SWER, label $ \in [0,2]$ & 1380 & \bf 51.5 & 49.9\\
Bottom 20\% SWER, label $ \in [3,5]$ & 2079 & \bf 43.0 & 42.2\\
\hline
\end{tabular}
\caption{\label{table:hardsts} Performance, measured in Pearson's $r \times 100$, for different data splits of the STS data. The first row shows performance across all unique examples, the next row shows the negation split, and the last four rows show difficult examples filtered symmetric word error rate (SWER). 
The last two rows show relatively easy examples according to SWER. 
}
\vspace{-4mm}
\end{table*}

It has been observed previously that the performance of Transformer models is sensitive to batch size~\cite{popel2018training}
. We found this to be especially true when training sequence-to-sequence models to learn sentence embeddings. Figure~\ref{fig:batchsize} shows plots of the average 2012-2016 STS performance of the learned sentence embedding as batch size increases for both the BiLSTM and Transformer. Initially, at a batch size of 2500 tokens, sentence embeddings learned are worse than random, even though validation perplexity does decrease during this time. Performance rises as batch size increases up to around 100,000 tokens. In contrast, the BiLSTM is more robust to batch size, peaking much earlier around 25,000 tokens, and even degrading at higher batch sizes.

\section{Model Ablations}
\label{appendix:ablations}

\begin{table}[h!]
\centering
\begin{tabular} { | l | c | c |}
\hline
Architecture & STS & Time (s) \\
\hline
Transformer (5L/1L) & 70.3 & 1767\\
Transformer (3L/1L) & 70.1 & 1548\\
Transformer (1L/1L) & 70.0 & 1244\\
Transformer (5L/5L) & 69.8 & 2799\\
\hline
\end{tabular}
\caption{\label{table:ablation} Results on the Semantic Textual Similarity (STS) datasets for different configurations of \engtrans, along with the time taken to train the model for 1 epoch. (XL/YL) means X layers were used in the encoder and Y layers in the decoder. Performance is measured in Pearson's $r \times 100$.
}
\end{table}

In this section, we vary the number of layers in the encoder and decoder in \trans. We see that performance increases as the number of encoder layers increases, and also that a large decoder hurts performance, allowing us to save training time by using a single layer. These results can be compared to those in Table~\ref{table:ablation} showing that Transformers outperform BiLSTMS in these experiments.

\section{Hard STS}
\label{appendix:hardsts}
We show further difficult splits in Table~\ref{table:hardsts}, including a negation split, beyond those used in {\it Hard STS} and compare the top two performing models in the STS task from Table~\ref{table:results}. We also show easier splits in the bottom of the table.

\end{document}